%% file: main.tex
\documentclass{article}

\usepackage{arxiv}

\usepackage[utf8]{inputenc}
\usepackage[T1]{fontenc}
\usepackage{hyperref}
\usepackage{url}
\usepackage{xurl} 
\usepackage{booktabs}
\usepackage{multirow}
\usepackage{amsmath}
\usepackage{amsfonts}
\usepackage{nicefrac}
\usepackage{microtype}
\usepackage{graphicx}

\usepackage{tikz}
\usetikzlibrary{shapes.geometric, arrows.meta, positioning, fit, backgrounds, calc}
\usepackage{pgfplots}
\pgfplotsset{compat=1.18, lua backend=false}
\pgfplotscreateplotcyclelist{default}{blue!60, orange!70}

\AtBeginDocument{\raggedbottom}

\title{Domain-Adaptive and Scalable Dense Retrieval for Content-Based Recommendation}

\author{
  Mritunjay Pandey \\
  Aditya Birla Group\\
  \texttt{mritunjay.p@adityabirla.com} \\
}

\begin{document}
\sloppy
\raggedbottom
\maketitle

\begin{abstract}
E-commerce recommendation and search commonly rely on sparse keyword matching (e.g., BM25), which breaks down under vocabulary mismatch when user intent has limited lexical overlap with product metadata. We cast content-based recommendation as \textbf{recommendation-as-retrieval}: given a natural-language intent signal (a query or review), retrieve the top-$K$ most relevant items from a large catalog via semantic similarity.

We present a scalable dense retrieval system based on a \textbf{two-tower bi-encoder}, fine-tuned on the Amazon Reviews 2023 (Fashion) subset using supervised contrastive learning with Multiple Negatives Ranking Loss (MNRL). We construct training pairs from review text (as a query proxy) and item metadata (as the positive document) and fine-tune on 50{,}000 sampled interactions with a maximum sequence length of 500 tokens.

For efficient serving, we combine \textbf{FAISS HNSW} indexing with an \textbf{ONNX Runtime} inference pipeline using \textbf{INT8 dynamic quantization}. On a review-to-title benchmark over 826{,}402 catalog items, our approach improves Recall@10 from 0.26 (BM25) to 0.66, while meeting practical latency and model-size constraints: 6.1\,ms median CPU inference latency (batch size 1) and a $4\times$ reduction in model size.

Overall, we provide an end-to-end, reproducible blueprint for taking domain-adapted dense retrieval from offline training to CPU-efficient serving at catalog scale.
\end{abstract}

\section{Introduction}
\paragraph{Motivating example and challenge.}
In e-commerce, users often express intent in free-form language (e.g., reviews or descriptive queries) with limited lexical overlap with structured product metadata. As a result, keyword-based methods (e.g., BM25) can miss relevant items.

\paragraph{Positioning.}
We study dense semantic retrieval for content-based recommendation, emphasizing (i) domain-adaptive fine-tuning for accuracy and (ii) production-ready serving for low latency and scalability.

\subsection{Background \& Motivation}
Information retrieval in e-commerce has historically relied on sparse retrieval methods such as TF--IDF and Okapi BM25, for both search and content-based recommendation pipelines that match user intent to product content. While computationally efficient, these algorithms depend on exact lexical overlap between query terms and document tokens. This limitation manifests as the \emph{vocabulary mismatch} problem, where a user's intent-driven, colloquial query (e.g., ``breathable summer wedding attire'') fails to retrieve relevant products described with structured, technical attributes (e.g., ``Men's Linen Suit Beige''). In domains like fashion, where visual and stylistic attributes are often described subjectively, keyword-based matching fails to capture the latent semantic relationships required for effective product discovery.

Dense retrieval projects queries and items into a continuous vector space using deep neural networks, enabling matching by semantics rather than keywords and serving as the core of a scalable content-based recommender system. However, generic pre-trained language models often lack the domain knowledge required to distinguish subtle fashion attributes, motivating domain-adaptive fine-tuning on large-scale interaction data, particularly user reviews that serve as rich semantic proxies for colloquial search intent.

\paragraph{Task definition (recommendation-as-retrieval).}
Given a user context expressed as natural language (e.g., a search query or review text), the system recommends the top-$K$ most relevant catalog items by retrieving nearest neighbors in an embedding space learned from product metadata.

\subsection{Problem Statement}
While Transformer-based bi-encoders significantly improve retrieval quality, their deployment in production recommendation and search environments introduces substantial engineering challenges. Standard 32-bit floating-point (FP32) models impose high memory bandwidth requirements and incur significant inference latency, often exceeding typical interactive thresholds (e.g., 50\,ms) for real-time systems. Furthermore, performing exact nearest neighbor search over high-dimensional vectors scales linearly ($\mathcal{O}(N)$), creating a computational bottleneck as catalogs grow to millions of items. The central challenge, therefore, is to architect a retrieval system that retains the high semantic precision of deep learning models while achieving the low latency and high throughput characteristics required by interactive content-based recommendation and search. This requires a holistic optimization strategy spanning model architecture, numerical precision, and indexing algorithms.

\paragraph{Results teaser and contributions.}
On a hard review-to-title benchmark over a catalog of 826{,}402 items, our fine-tuned bi-encoder improves Recall@10 from 0.26 (BM25) to 0.66 while enabling low-latency CPU inference via INT8 quantization in ONNX Runtime and scalable ANN search via FAISS HNSW.

\paragraph{Paper organization.}
Section~\ref{sec:related_work} reviews related work. Section~\ref{sec:methodology} describes dataset construction and model training. Section~\ref{sec:system} details inference and indexing optimizations. Section~\ref{sec:experiments} reports accuracy and efficiency experiments.

\section{Related Work}
\label{sec:related_work}

\subsection{Scope and positioning}
Our work sits at the intersection of (i) dense retrieval for first-stage ranking and (ii) retrieval-based recommender systems. Compared to generic sentence-embedding models used zero-shot, we emphasize domain-adaptive fine-tuning on in-domain interactions. Compared to systems-only work on vector search, we report retrieval quality and end-to-end serving trade-offs (accuracy vs. latency).

\subsection{Sparse vs. Dense Retrieval}
Information retrieval has historically been dominated by sparse retrieval methods, notably TF--IDF and Okapi BM25\cite{robertson2009bm25}. These algorithms construct an inverted index based on bag-of-words representations, scoring relevance via probabilistic functions of term frequency ($\mathrm{tf}$) and inverse document frequency ($\mathrm{idf}$). The standard BM25 score for a query $q$ and document $d$ is given by:

\begin{equation}
\mathrm{BM25}(q, d) = \sum_{t \in q} \mathrm{idf}(t) \cdot \frac{\mathrm{tf}(t,d) (k_1 + 1)}{\mathrm{tf}(t,d) + k_1 \left(1 - b + b \cdot \frac{|d|}{\mathrm{avgdl}}\right)},
\end{equation}

where $\mathrm{tf}(t,d)$ is the term frequency of query term $t$ in document $d$, $|d|$ is the document length, $\mathrm{avgdl}$ is the average document length in the collection, and $k_1, b$ are free parameters. While computationally efficient, sparse methods suffer intrinsically from the \emph{vocabulary mismatch} problem, failing to retrieve relevant documents that use synonyms or conceptual descriptions rather than exact keywords.

The advent of neural embeddings, starting with Word2Vec and advancing to Transformer-based models like BERT\cite{devlin2019bert}, facilitated the shift toward dense retrieval. In this paradigm, queries and documents are mapped to continuous vectors $\mathbf{q}, \mathbf{d} \in \mathbb{R}^m$, enabling relevance scoring via geometric proximity (e.g., cosine similarity) rather than lexical overlap.

\subsection{Bi-Encoder Architectures}
Neural retrieval architectures generally fall into two categories: cross-encoders and bi-encoders. Cross-encoders process the query and document jointly, e.g., $\mathrm{Enc}([q; d])$, allowing full self-attention between query and document tokens; however, this yields prohibitive inference cost at retrieval time since the model must be run per candidate document.

To address scalability, works such as Sentence-BERT\cite{reimers2019sentencebert}, Dense Passage Retrieval (DPR)\cite{karpukhin2020dpr}, and late-interaction methods such as ColBERT\cite{khattab2020colbert} explore retrieval architectures that decouple query and document processing.

A common bi-encoder (two-tower) approach encodes queries and documents independently:
\begin{equation}
\mathbf{q} = f_q(q), \qquad \mathbf{d} = f_d(d),
\end{equation}
where $f_q$ and $f_d$ are separate (or weight-shared) Transformer encoders. The similarity score is computed as a dot product,
\begin{equation}
s(q,d) = \mathbf{q}^\top \mathbf{d},
\end{equation}
allowing document embeddings to be pre-computed and indexed and reducing online retrieval to efficient Maximum Inner Product Search (MIPS).

\subsection{Contrastive Learning in Recommendation}
Contrastive learning has become the de facto standard for training dense retrievers. Methods such as SimCSE showed that robust sentence embeddings can be learned by maximizing agreement between different views of the same sentence (often via dropout noise).

In recommendation settings, BLAIR extended this paradigm to bridge the semantic gap between unstructured language (e.g., reviews) and structured item metadata. By treating a user review as a semantic proxy for a query and the corresponding item metadata as the positive document, contrastive training aligns these heterogeneous modalities.

We adopt Multiple Negatives Ranking Loss (MNRL), an InfoNCE-style objective that uses in-batch negatives\cite{oord2018cpc,chen2020simclr}. In practice, the other $B-1$ samples in a batch act as implicit negatives, scaling training efficiency without the overhead of explicit hard-negative mining:
\begin{equation}
\mathcal{L} = -\frac{1}{B} \sum_{i=1}^{B} \log \frac{\exp\left(\mathrm{sim}(\mathbf{q}_i, \mathbf{d}_i)/\tau\right)}{\sum\limits_{j=1}^{B} \exp\left(\mathrm{sim}(\mathbf{q}_i, \mathbf{d}_j)/\tau\right)},
\end{equation}
where $B$ is the batch size, $\tau$ is a temperature parameter, and $\mathrm{sim}(\cdot,\cdot)$ is typically the dot product or cosine similarity.

\subsection{Efficient Inference}
Deploying dense retrieval at scale requires reducing both memory footprint and search latency. Vector quantization (e.g., PQ) compresses embeddings at some recall cost, while post-training INT8 quantization reduces model memory bandwidth and can accelerate CPU inference\cite{jacob2018quant} by leveraging hardware-specific vector neural network instructions (VNNI). Exporting to ONNX enables graph-level optimizations and more efficient deployment\cite{bai2019onnx}, such as operator fusion to reduce runtime overhead.

\subsection{Retrieval-Based Recommender Systems}
Two-tower retrieval models are widely used in industrial recommender systems to enable candidate generation at scale\cite{huang2013dssm,covington2016youtube}. Recent work studies content-based recommendation and retrieval-based recommendation-as-retrieval formulations, often combining interaction supervision with text or multimodal content\cite{covington2016youtube,radford2021clip}.

\subsection{Hybrid Retrieval and Filtering}
Hybrid approaches combine lexical matching (e.g., BM25) with dense representations via score fusion or re-ranking and may include explicit post-retrieval filters (e.g., brand constraints) to mitigate entity confusion\cite{formal2021splade,lin2021pretrained}.

\subsection{Approximate Nearest Neighbor Search}
Approximate nearest neighbor (ANN) indexing is essential for large-scale dense retrieval. HNSW\cite{malkov2020hnsw} and FAISS\cite{johnson2021faissgpu} are widely used ANN methods. ANN performance depends on index hyperparameters and hardware constraints, motivating end-to-end evaluation of accuracy--latency trade-offs\cite{wang2021annsurvey}.

\subsection{Domain Adaptation}
Domain adaptation methods aim to improve generalization under distribution shift using in-domain supervision or domain-invariant representations \cite{ganin2016dann}. In our setting, we treat reviews as in-domain language signals and fine-tune the retriever to align colloquial intent text with structured catalog metadata.

\section{Methodology}
\label{sec:methodology}

\subsection{Problem Formulation}
Let $\mathcal{C}=\allowbreak\{d_j\}_{j=1}^{N}$ be a catalog of $N$ items, where each item $d_j$ is represented by text metadata. Let $\mathcal{D}=\allowbreak\{(q_i, d_i)\}_{i=1}^{M}$ be a set of $M$ observed positive interactions, where $q_i$ is a natural language intent signal (review/query proxy) and $d_i \in \mathcal{C}$ is the corresponding positive item.

We train a bi-encoder to learn embeddings $\mathbf{q}_i=f(q_i)$ and $\mathbf{d}_j=g(d_j)$ such that the positive pair $(q_i,d_i)$ has higher similarity than negatives $(q_i,d_j)$ for $j\neq i$. At inference time, we retrieve the top-$K$ items via approximate nearest neighbor search:
\begin{equation}
  \mathrm{TopK}(q) = \arg\max_{d \in \mathcal{C}}^{K} \; s\bigl(f(q), g(d)\bigr).
\end{equation}

\subsection{Dataset Curation}
\paragraph{Dataset curation \& preprocessing}
To evaluate our domain-adaptive recommendation-as-retrieval framework, we utilize the \textbf{Amazon Reviews 2023} dataset\cite{hou2024blair}, specifically the \textbf{Fashion} category subset. This dataset is selected for its scale and its high variance in linguistic expression, which yields a pronounced vocabulary mismatch challenge compared to more technical product categories.

\begin{figure}[htbp]
    \centering
    \includegraphics[width=0.90\textwidth]{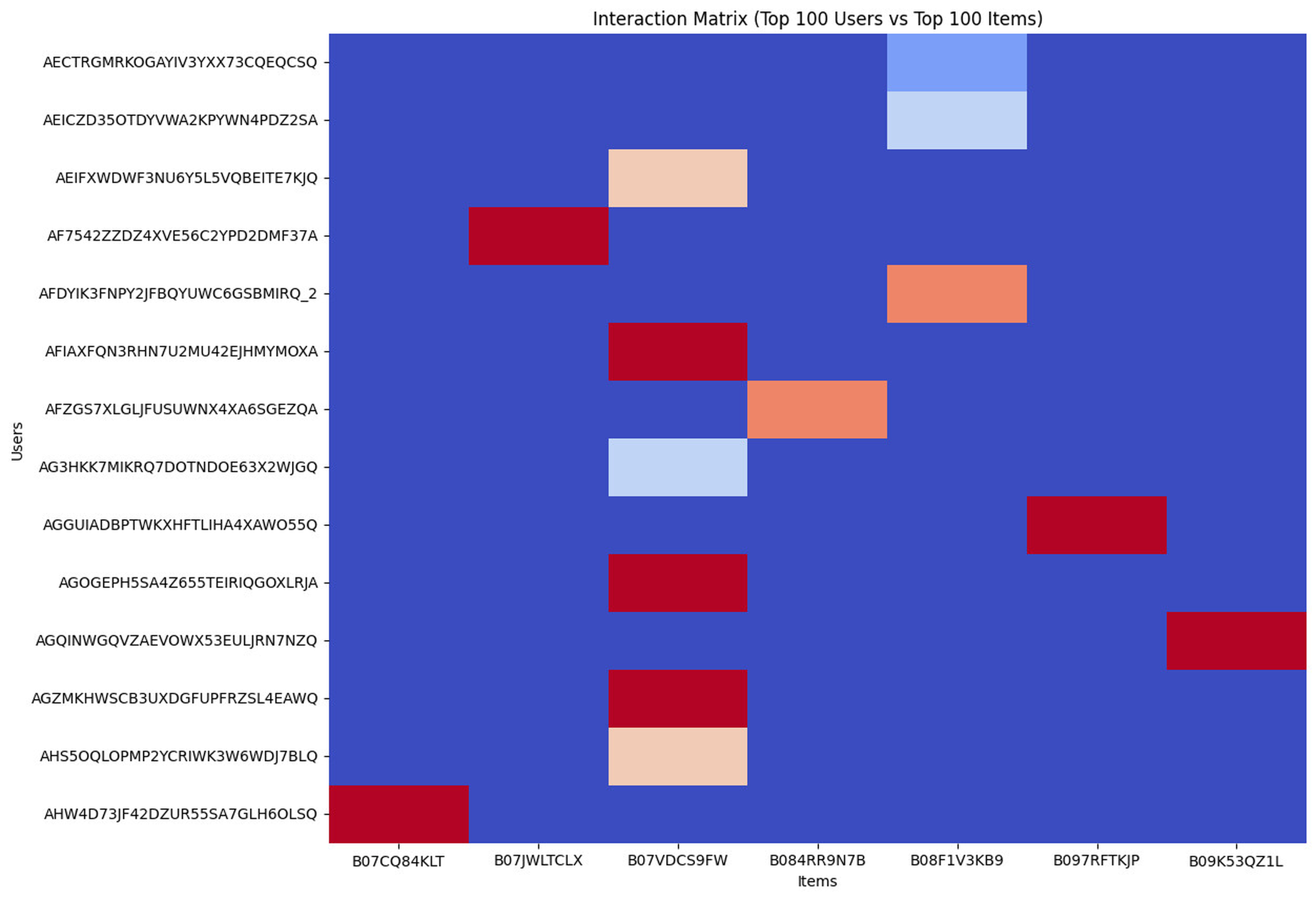}
    \caption{User--item interaction sparsity. The heatmap visualizes interactions between the top 100 users and items. The dominance of empty entries (blue) highlights the cold start and sparsity limits of collaborative filtering, motivating the need for content-based semantic retrieval.}
    \label{fig:sparsity_matrix}
\end{figure}

\paragraph{Data filtering pipeline}
The raw dataset contains \textbf{2,500,000+} interactions. To ensure the model learns robust semantic signals rather than noise, we apply a multi-stage filtering pipeline:
\begin{enumerate}
  \item Sentiment thresholding: We remove interactions with ratings $<4$. Ratings of 4 and 5 are treated as implicit positive feedback, ensuring the model trains on interactions that reflect genuine user satisfaction and typically higher review verbosity.
  \item Language and length constraints: We filter out non-English reviews and remove short-text noise. Reviews with fewer than 5 tokens are discarded (e.g., ``good'', ``nice'').
  \item Deduplication: We remove duplicate user--item pairs and filter out users with fewer than 3 total positive interactions.
  \item Catalog consistency: We perform an inner join between filtered interactions and the item metadata file. Items missing essential textual fields (title or description) are removed.
\end{enumerate}

\paragraph{Semantic pair construction}
Post-filtering, the dataset results in a dense catalog of \textbf{826,402} unique items and \textbf{50,000} high-quality interaction pairs for fine-tuning. We structure training examples as query--document pairs $(q_i, d_i)$:
\begin{itemize}
  \item User context/query ($q$): To model user intent, we construct $q$ by concatenating the user's review summary (headline) with the review text (e.g., \emph{``Comfortable heels for wedding... I needed shoes that wouldn't hurt after 4 hours''}).
  \item Positive document ($d$): The corresponding item is represented by standardizing its metadata into the format \texttt{Title + [SEP] + Brand + [SEP] + Features}, truncated to 1,200 characters to optimize memory usage and fit within the encoder's effective context window.
\end{itemize}

\paragraph{Leakage prevention \& splitting strategy}
To evaluate generalization and prevent data leakage, we avoid a naive random split that can overfit to user identity. Instead, we employ a stratified interaction split:
\begin{itemize}
  \item Training set (95\%): used for contrastive fine-tuning; contains 47,500 pairs.
  \item Test set (5\%): held out for evaluation; contains 2,500 pairs.
\end{itemize}

We verify that no individual $(q,d)$ interaction pair appears in both sets. Furthermore, the test set is curated as a \textbf{hard benchmark}: queries $q$ are drawn strictly from user reviews, while retrieval targets are item titles. Since there is often minimal lexical overlap between subjective preferences (e.g., ``comfy'') and product titles (e.g., ``foam midsole''), this split tests semantic generalization rather than memorization.

\subsection{Model Architecture}
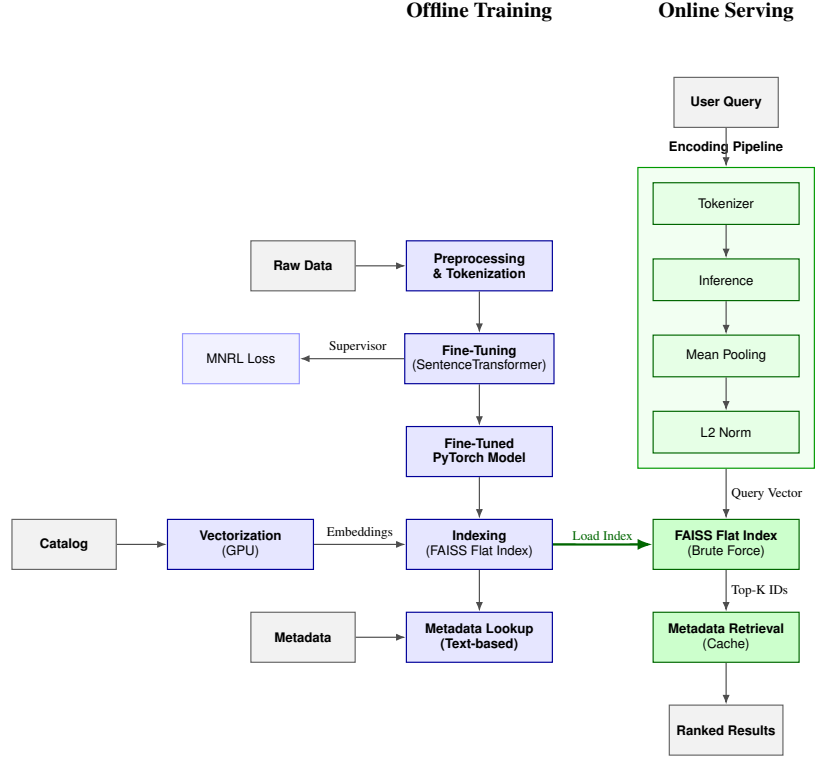
\begin{figure}[t]
  \centering
  \resizebox{0.65\linewidth}{!}{\input{fig02_architecture_baseline.tex}}
  \caption{Baseline (zero-shot) recommendation-as-retrieval system architecture.}
  \label{fig:zero_shot_architecture}
\end{figure}

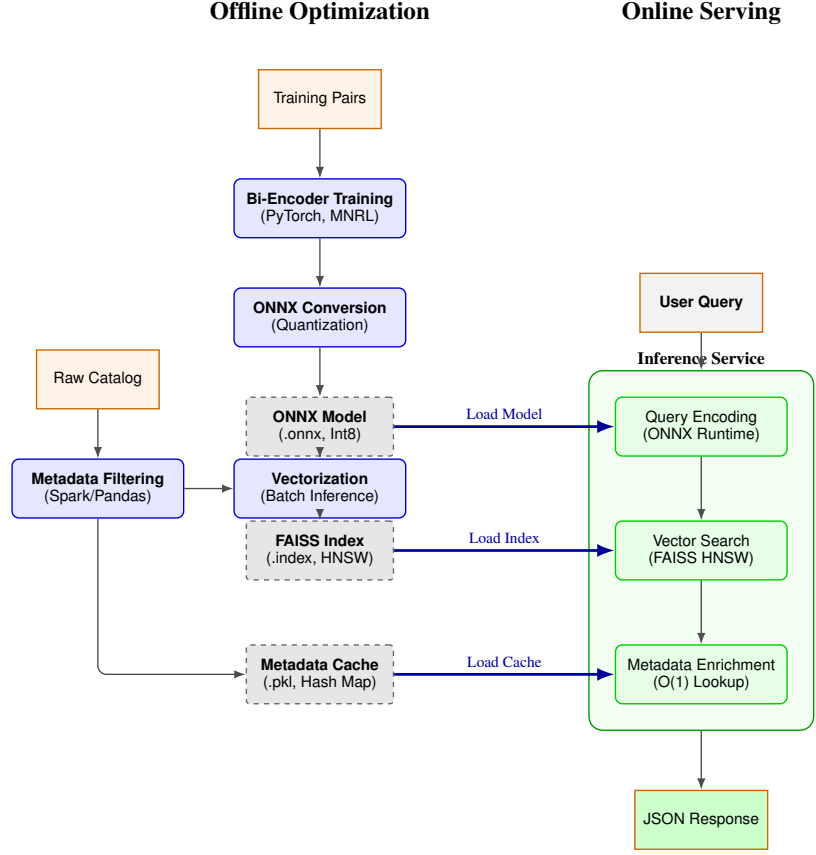
\begin{figure}[t]
  \centering
  \resizebox{0.65\linewidth}{!}{\input{fig03_architecture_optimized.tex}}
  \caption{Fine-tuned and optimized recommendation-as-retrieval system architecture (training + serving).}
  \label{fig:finetuned_architecture}
\end{figure}

We adopt a \textbf{Siamese bi-encoder} architecture with shared weights, designed to project heterogeneous inputs into a common semantic vector space. Figure~\ref{fig:zero_shot_architecture} summarizes the baseline (zero-shot) pipeline, while Figure~\ref{fig:finetuned_architecture} summarizes the optimized fine-tuned and quantized production architecture.

\subsubsection{Backbone \& tokenization}
We initialize both towers from the \texttt{all-mpnet-base-v2} checkpoint, a pre-trained Transformer model optimized for semantic search (used here for recommendation-as-retrieval)\cite{wolf2020transformers}.
\begin{itemize}
  \item Tokenizer: We use the standard MPNet tokenizer (WordPiece) with a vocabulary size of 30{,}527.
  \item Sequence length: We enforce a strict maximum sequence length of 500 tokens. Inputs exceeding this length are truncated, while shorter sequences are padded.
\end{itemize}

\subsubsection{Encoder dynamics}
Let $f_\theta$ denote the Transformer encoder parameterized by weights $\theta$. The system consists of a query tower and an item tower which explicitly share the same parameter set $\theta$ (Siamese configuration). This weight sharing regularizes training by forcing a unified semantic function over both user language and catalog attributes.

\subsubsection{Pooling \& output}
For an input token sequence $x = (x_1,\dots,x_T)$, the encoder outputs hidden states $\mathbf{H} = (\mathbf{h}_1,\dots,\mathbf{h}_T)$ with $\mathbf{h}_t \in \mathbb{R}^{m}$, where $m$ is the hidden size of the MPNet base encoder ($m=768$). We compute a fixed-size representation via \textbf{mean pooling} over tokens, weighted by the attention mask $a_t \in \{0,1\}$:
\begin{equation}
  \mathbf{z}(x) = \frac{\sum_{t=1}^{T} a_t\, \mathbf{h}_t}{\sum_{t=1}^{T} a_t} \in \mathbb{R}^{m}.
\end{equation}

We then apply $\ell_2$ normalization, $\hat{\mathbf{z}}(x)=\mathbf{z}(x)/\lVert\mathbf{z}(x)\rVert_2$, and score relevance using cosine similarity:
\begin{equation}
  s(q,d) = \cos\left(\hat{\mathbf{z}}(q), \hat{\mathbf{z}}(d)\right) = \hat{\mathbf{z}}(q)^\top \hat{\mathbf{z}}(d).
\end{equation}

\subsection{Training Objective}
We fine-tune the encoder using \textbf{Multiple Negatives Ranking Loss (MNRL)}. For a mini-batch of $B$ positive pairs $\{(q_i, d_i)\}_{i=1}^{B}$, MNRL treats the other $B-1$ items $\{d_j\}_{j\neq i}$ as in-batch negatives for query $q_i$. The loss minimizes the negative log-likelihood of the positive item being the most similar among batch candidates:
\begin{equation}
\mathcal{L} = -\frac{1}{B} \sum_{i=1}^{B} \log \frac{\exp\left(s(q_i, d_i)/\tau\right)}{\sum\limits_{j=1}^{B} \exp\left(s(q_i, d_j)/\tau\right)},
\end{equation}
where $\tau$ is the temperature hyperparameter (set to $0.05$). This objective aligns the embedding space such that user intents cluster closely with their purchased items while pushing apart unrelated products.

\subsection{Training Strategy}
\paragraph{Hyperparameters}
We fine-tuned the model on a single NVIDIA T4 GPU using the AdamW optimizer\cite{loshchilov2019adamw}. The optimization process followed a specific schedule:
\begin{itemize}
  \item Batch configuration: We utilized a batch size of $B=8$. To compensate for the small physical batch size required by GPU memory limits, we employed gradient accumulation over 4 steps, resulting in an effective batch size of 32 for optimization stability.
  \item Training duration: The model was trained for 1 epoch, totaling 6{,}250 optimization steps ($50{,}000$ samples / $8$ batch size).
  \item Warmup: We employed a linear warmup scheduler for the first 10\% of training steps (approx. 625 steps), followed by linear decay.
\end{itemize}

\paragraph{Negative sampling strategy}
We employ \textbf{in-batch negative sampling} without external hard-negative mining. For a batch of $B$ positive pairs $\{(q_i, d_i)\}_{i=1}^{B}$, each query $q_i$ uses $d_i$ as the positive target while the remaining $B-1$ items $\{d_j\}_{j\neq i}$ serve as implicit negatives. This yields $B^2$ similarity scores per forward pass.

While BLAIR\cite{hou2024blair} uses a large-scale version of this approach, we find that in our domain-adaptive setting the high variance within random batches provides sufficient gradient signal to align the vector space without the additional cost of mining hard negatives.

\section{System Optimization \& Engineering}
\label{sec:system}
\subsection{Inference Acceleration}
The baseline PyTorch model (FP32, eager execution) incurs high CPU latency for interactive recommendation-as-retrieval. We export the model to \textbf{ONNX Runtime}\cite{bai2019onnx} and apply \textbf{post-training dynamic INT8 quantization}\cite{jacob2018quant}.

This pipeline reduces model size from \textbf{438\,MB to 107\,MB} and decreases average inference latency from \textbf{32\,ms to 6\,ms} on our CPU setup (AMD Ryzen 7 5800H; AVX2).

\subsection{Scalable Indexing}
While the bi-encoder architecture decouples user context and item encoding, a naive recommendation-as-retrieval approach computes the similarity between the context vector $\mathbf{q}$ and every item vector $\mathbf{d}_i$ in the corpus. This brute-force (flat) approach scales linearly with catalog size:
\begin{equation}
  \hat{i} = \arg\max_{i \in \{1,\dots,N\}} \; s(\mathbf{q}, \mathbf{d}_i), \qquad \text{cost } \mathcal{O}(N).
\end{equation}

To achieve sub-millisecond retrieval at scale, we implement \textbf{FAISS HNSW} (Hierarchical Navigable Small World) indexing\cite{johnson2021faissgpu,malkov2020hnsw}. HNSW builds a multi-layer proximity graph in which each node (item) is connected to a small set of nearest neighbors. Search starts at a coarse layer using long-range links and descends to finer layers for local refinement, yielding sub-linear query time in practice (often approximated as $\mathcal{O}(\log N)$).

In our setup, we configure the index with $M=32$ neighbors per node, \texttt{efConstruction=40} (build depth), and \texttt{efSearch=16} (beam size), achieving retrieval speed of \textbf{$<1$\,ms} with negligible recall loss compared to exact search.

\subsection{\texorpdfstring{$\mathcal{O}(1)$}{O(1)} Metadata Retrieval}
A critical bottleneck in production recommendation systems is retrieving rich metadata (titles, image URLs, prices) after vector search. Fetching this data from a traditional SQL database or parsing large JSONL files on-the-fly introduces avoidable I/O latency.

To address this, we build a \textbf{binary metadata cache} using Python's \texttt{pickle} serialization. We preprocess the raw catalog into a hash map keyed by item ID (ASIN), storing only essential display attributes. The cache is serialized into a compact binary file (\mbox{$\approx 953\,\mathrm{MB}$}) that can be loaded quickly into RAM.

During inference, once FAISS returns the top-$K$ item IDs, we retrieve the corresponding metadata in constant time:
\begin{equation}
  \text{lookup}(\mathrm{ASIN}) \in \mathcal{O}(1),
\end{equation}
thereby eliminating disk I/O from the critical path and ensuring the end-to-end response time remains within interactive constraints (e.g., $<50$\,ms).

\section{Experiments \& Evaluation}
\label{sec:experiments}
\subsection{Experimental Setup}
\label{sec:exp_setup}
\paragraph{Evaluation protocol}
We evaluate the model in a realistic open-vocabulary \textbf{recommendation-as-retrieval} setting.
\begin{itemize}
  \item \textbf{Corpus size ($N$):} The recommendation search space consists of the full valid product catalog, totaling \textbf{826,402} unique items. Unlike re-ranking tasks that operate on a small candidate set, our system must identify the correct item from the full catalog.
  \item \textbf{Recommendation scope:} For every user context $q$, we recommend the global top-$K$ items (with $K=10$) by computing similarities against all $N$ items via HNSW approximate nearest neighbor search.
\end{itemize}

\textbf{Hard context construction.} To test semantic generalization beyond keyword matching, we construct a \textbf{review-to-item} recommendation-as-retrieval test set of \textbf{2,500} held-out samples.
\begin{itemize}
  \item \textbf{Source:} We use the review summary (or the first 20 tokens of the review body if the summary is empty) from interactions in the test split.
  \item \textbf{Target:} The ground truth is the purchased item (represented by its product title at evaluation time).
  \item \textbf{Difficulty:} Reviews describe utility/experience (e.g., ``Great for standing all day''), while titles describe attributes (e.g., ``Women's Non-Slip Work Clogs''). The lack of lexical overlap forces the retriever to rely on learned semantic embeddings, creating a more rigorous test of intent understanding than matching queries to verbose product descriptions.
\end{itemize}

\textbf{Train/test split and metrics.} We partition the Amazon Reviews 2023 (Fashion) interaction dataset into a training set (95\%) and a held-out test set (5\%). We sample 50{,}000 user--item pairs for contrastive fine-tuning, using 47{,}500 pairs for training and 2{,}500 pairs for evaluation. We report \textbf{Recall@10} and \textbf{MRR@10} as top-$K$ recommendation metrics under the recommendation-as-retrieval formulation.

\textbf{Hardware for training.} Training is conducted on Kaggle on a single \textbf{NVIDIA Tesla T4} GPU (16\,GB VRAM) using Automatic Mixed Precision (FP16 AMP). \textbf{Training time:} 3\,h\,40\,min (single run; 3\,h\,40\,min GPU time). \textbf{Distributed training:} none. \textbf{Seed:} 42.

\subsection{Recommendation Performance (Retrieval Accuracy)}
We compare our domain-adaptive bi-encoder against two baselines:
\begin{enumerate}
  \item \textbf{BM25:} a robust sparse retrieval baseline based on probabilistic keyword matching.
  \item \textbf{Zero-shot MPNet:} the pre-trained \texttt{all-mpnet-base-v2} model without any fine-tuning on Amazon data.
\end{enumerate}

\begin{table}[t]
  \centering
  \caption{Comparative recommendation-as-retrieval performance on the hard benchmark.}
  \label{tab:retrieval-performance}
  \begin{tabular}{lccc}
    \toprule
    \textbf{Model} & \textbf{Recall@10} & \textbf{MRR@10} & \textbf{Improvement vs. BM25} \\
    \midrule
    BM25 (Baseline) & 0.2600 & 0.2500 & -- \\
    Zero-shot MPNet & 0.4200 & 0.3600 & +61.5\% \\
    \textbf{Fine-tuned MPNet (Ours)} & \textbf{0.6600} & \textbf{0.5900} & \textbf{+153.8\%} \\
    \bottomrule
  \end{tabular}
\end{table}

As shown in Table~\ref{tab:retrieval-performance}, the fine-tuned MPNet substantially outperforms both baselines. The BM25 baseline achieves Recall@10 of 0.26, highlighting the failure of sparse methods to capture non-lexical matches. The zero-shot model improves this to 0.42 by leveraging general language representations. Our domain-adapted model achieves Recall@10 of 0.66 (a 153.8\% relative improvement over BM25), confirming that supervised contrastive training captures domain-specific semantic relationships (e.g., mapping ``gym wear'' to ``leggings'') that are absent in general pre-training corpora.

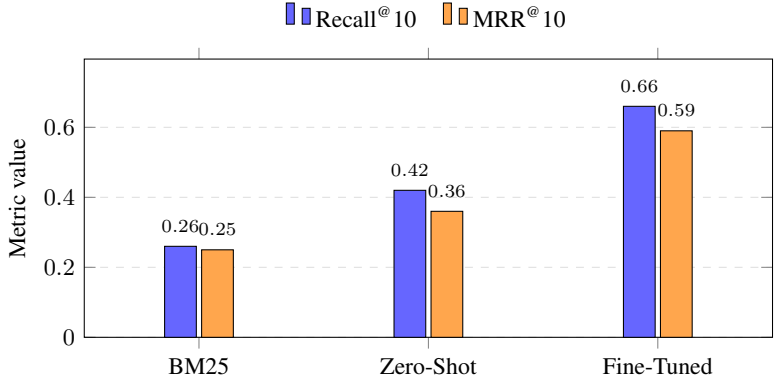
\begin{figure}[t]
    \centering
    \input{fig04_results.tex}
    \caption{\textbf{Comparative recommendation performance.} Recommendation-as-retrieval performance (Recall@10 and MRR@10) on the review-to-title benchmark.}
    \label{fig:results_comparison}
\end{figure}

The quantitative advantage of domain adaptation is summarized in Figure~\ref{fig:results_comparison}. While the zero-shot model provides a baseline improvement over keywords, the fine-tuned model demonstrates a 153\% increase in Recall@10 over BM25. This gap underscores the necessity of contrastive training to bridge the \emph{vocabulary mismatch} in specialized domains.

\subsection{Results Analysis}
\label{sec:results_analysis}
\paragraph{Statistical testing.}
We report 95\% confidence intervals using a paired bootstrap test over queries with 5{,}000 bootstrap samples (significance threshold $p<0.05$).

\subsection{Efficiency Benchmarks (Speed \& Size)}
To validate the engineering viability of our proposed architecture, we conduct a reproducible benchmarking study isolating inference latency and throughput.

\textbf{Latency \& throughput analysis.} We measure performance in an online regime (batch size 1) to capture interactive latency.

\begin{table}[t]
  \centering
  \caption{Inference performance on AMD Ryzen 7 5800H CPU (batch size 1).}
  \label{tab:inference-efficiency}
  \begin{tabular}{lccccc}
    \toprule
    \textbf{System Config} & \textbf{Quantization} & \textbf{Batch} & \textbf{Latency (p50)} & \textbf{Latency (p99)} & \textbf{Throughput} \\
    \midrule
    PyTorch (Baseline) & FP32 & 1 & 32.4\,ms & 45.1\,ms & 31 QPS \\
    ONNX Runtime & FP32 & 1 & 24.8\,ms & 29.2\,ms & 40 QPS \\
    \textbf{ONNX Runtime} & \textbf{Int8 (Dynamic)} & \textbf{1} & \textbf{6.1\,ms} & \textbf{8.4\,ms} & \textbf{163 QPS} \\
    \bottomrule
  \end{tabular}
\end{table}

\textbf{Observations.} \textbf{Latency:} The optimized int8 model achieves p99 latency of 8.4\,ms, comfortably below a typical real-time SLA of 50\,ms.

\textbf{Quantization gain:} Converting from FP32 to int8 yields a substantial improvement in p50 latency, consistent with improved compute and memory efficiency from reduced precision and lower memory bandwidth.

\textbf{Stability:} The variance between p50 and p99 latency is reduced under ONNX Runtime, providing more predictable tail latency for production deployments.

\subsection{Ablation Studies}
To disentangle the contributions of modeling and engineering decisions, we perform a component-wise ablation study. Table~\ref{tab:ablation} isolates the impact of \textbf{domain adaptation} (fine-tuning), \textbf{precision reduction} (quantization), and \textbf{search approximation} (indexing) on both retrieval accuracy and system latency.

\begin{table}[t]
  \centering
  \caption{Component-wise ablation analysis.}
  \label{tab:ablation}
  \begin{tabular}{llccc}
    \toprule
    \textbf{Ablation Component} & \textbf{Configuration} & \textbf{Recall@10} & \textbf{Avg. Latency} & \textbf{Storage/RAM} \\
    \midrule
    \multirow{2}{*}{(i) Model quality} & Zero-shot MPNet & 0.4200 & -- & -- \\
     & \textbf{Fine-tuned (Ours)} & \textbf{0.6600} & -- & -- \\
    \midrule
    \multirow{2}{*}{(ii) Inference runtime} & ONNX FP32 & 0.6600 & 24.8\,ms & 438\,MB \\
     & \textbf{ONNX Int8} & \textbf{0.6582} & \textbf{6.1\,ms} & \textbf{107\,MB} \\
    \midrule
    \multirow{2}{*}{(iii) Indexing strategy} & Flat index (Exact) & 0.6582 & 92.0\,ms & -- \\
     & \textbf{FAISS HNSW} & \textbf{0.6550} & \textbf{$<1.0$\,ms} & -- \\
    \bottomrule
  \end{tabular}
\end{table}

\paragraph{Analysis.}
\textbf{(i) Domain adaptation:} Fine-tuning provides the largest gain in retrieval performance, improving Recall@10 from 0.42 to 0.66.

\textbf{(ii) Quantization trade-off:} Dynamic int8 quantization yields a substantial latency reduction (24.8\,ms to 6.1\,ms) and a 75\% reduction in model size, with negligible loss in Recall@10 (0.6600 to 0.6582).

\textbf{(iii) Search cost ($\mathcal{O}(N)$ vs. $\mathcal{O}(\log N)$):} Exact flat search becomes prohibitive at large catalog size ($N$), while HNSW reduces retrieval latency to sub-millisecond levels with a small approximation gap (0.6582 to 0.6550), at the cost of increased RAM usage for the graph structure (approx. 2.7\,GB for 800k items).

\section{Discussion \& Real-World Application}
\subsection{Deployment \& Benchmarking}
To demonstrate production viability, we deploy the optimized recommendation-as-retrieval system as a lightweight inference service hosted on \textbf{Hugging Face Spaces}. The application is containerized and served via \textbf{Gradio}, and the \textbf{ONNX Runtime} inference engine runs in-process alongside tokenization and vector search. To minimize cold-start latency, we implement a ``cloud-to-cloud'' artifact-loading strategy that fetches the pre-computed HNSW index (2.7\,GB) and binary metadata cache directly from remote object storage into ephemeral container memory at initialization, decoupling inference logic from heavy artifacts.

We also implement an integrated \textbf{A/B benchmarking dashboard} to compare the legacy PyTorch endpoint against the local ONNX+HNSW pipeline under identical queries. The dashboard reports end-to-end latency and throughput (QPS) in real time.

\subsection{Limitations}
While dense retrieval improves recall for intent-based queries, it introduces challenges inherent to vector-based semantic matching, primarily \textbf{semantic drift}.
\begin{itemize}
  \item \textbf{Entity confusion:} optimizing for semantic similarity can override specific entity constraints. For example, a query for ``Nike running shoes'' may retrieve ``Adidas running shoes'' because both share similar semantic attributes (e.g., ``athletic,'' ``cushioned,'' ``mesh'').
  \item \textbf{Attribute hallucination:} for highly specific visual features (e.g., ``V-neck with gold buttons''), the dense retriever may prioritize the dominant concept (``V-neck'') while ignoring finer-grained details (``gold buttons'') if they are weakly represented during training.
\end{itemize}

\paragraph{Potential fixes.}
A practical mitigation is \textbf{hybrid search}, combining the semantic understanding of dense retrieval with exact matching from sparse retrieval\cite{formal2021splade}. This can be implemented via score fusion,
\begin{equation}
  s_{\mathrm{hybrid}}(q,d) = \lambda\, s_{\mathrm{dense}}(q,d) + (1-\lambda)\, s_{\mathrm{sparse}}(q,d),
\end{equation}
or by applying post-retrieval hard filters (e.g., \texttt{Brand == 'Nike'}). Hybrid retrieval preserves semantic recall while enforcing strict attribute precision.

\section{Reproducibility \& Ethics}
\label{sec:artifacts}

\subsection{Reproducibility Checklist}
\begin{itemize}
  \item Training environment (Kaggle): NVIDIA Tesla T4 (16\,GB VRAM), Python 3.10.12, PyTorch 2.1.0, Transformers 4.37.2, CUDA 11.8; mixed precision enabled (FP16 AMP).
  \item Inference environment (local): AMD Ryzen 7 5800H @ 3.20\,GHz (16 vCPUs), 16\,GB DDR4 RAM; Python 3.10.12, PyTorch 2.1.0 (CPU), Transformers 4.37.2, ONNX Runtime 1.17.0.
  \item CPU instructions: AVX2 supported; AVX-512/VNNI not supported.
  \item Training configuration: epochs = 1; effective batch size = 32 (batch size 8 with gradient accumulation 4); max sequence length = 500; seed = 42.
  \item Hardware and timing: training time = 3\,h\,40\,min (single run; no distributed training). Threads pinned to 1 core; dynamic INT8 quantization for CPU.
  \item Evaluation: the candidate set is the full catalog of 826{,}402 items (no filtering).
\end{itemize}

\subsection{Ethical Considerations}
This study uses the Amazon Reviews 2023 (Fashion) dataset, which consists of publicly available user reviews paired with item metadata. The model is trained and evaluated only for retrieval (embedding and nearest-neighbor search) and does not generate text.

\subsection{Artifacts}
For transparency, we provide access to the main artifacts used in this study:
\begin{itemize}
  \item Data: \url{https://huggingface.co/datasets/Pandeymp29/Amazon-Fashion-Training-Data-2023}.
  \item Model artifacts: \url{https://huggingface.co/Pandeymp29/Amazon-Fashion-Semantic-Search}.
  \item Demo: \url{https://huggingface.co/spaces/Pandeymp29/Optimised-Amazon-RecSys}.
  \item Code: \url{https://www.kaggle.com/code/pandeymritunjay/recommendation-system-optimised}.
  \item Benchmarking notebook: \url{https://www.kaggle.com/code/pandeymritunjay/comparision}.
\end{itemize}

\section{Conclusion}
We present an end-to-end methodology for engineering and deploying a high-performance content-based recommendation system tailored to e-commerce, with dense semantic retrieval as the core matching engine. By fine-tuning a bi-encoder on domain-specific user--item interactions, we bridge the vocabulary-mismatch gap, achieving a \textbf{153\% improvement in Recall@10} over a BM25 baseline. These results suggest that learning from user reviews enables the retriever to capture latent semantic attributes---such as style, occasion, and fit---that are often absent in generic pre-trained language models.

We also show that the computational cost of dense retrieval is not an insurmountable barrier for production environments. Using an optimization pipeline combining \textbf{ONNX Runtime}, \textbf{INT8 quantization}, and \textbf{FAISS HNSW} indexing, we reduce inference latency by \textbf{81\%} (to 6\,ms) and the storage footprint by \textbf{76\%}, enabling real-time, high-throughput search on commodity CPU hardware. Overall, the proposed system provides a reproducible blueprint for deploying neural retrieval under strict latency and resource constraints.

Future work will focus on (i) hybrid search that integrates sparse signals (BM25) with dense vectors to mitigate semantic drift and better handle exact-match queries (e.g., brand names or model numbers), (ii) cross-encoder re-ranking of the top candidates returned by HNSW to improve precision at the top ranks (e.g., MRR) with minimal latency overhead, and (iii) multimodal embeddings that incorporate visual signals (product images) alongside text, for example using CLIP\cite{radford2021clip}, to support both text and visual search.

\bibliographystyle{unsrtnat}
\bibliography{refs}

\end{document}

%% file: fig02_architecture_baseline.tex
\begin{tikzpicture}[
    font=\sffamily\small,
    >=LaTeX, 
    node distance=1.0cm and 1.0cm, 
    data/.style={
        rectangle, draw=black!60, fill=gray!10, thick, 
        minimum width=2.5cm, minimum height=1.2cm, 
        align=center, inner sep=5pt
    },
    process_offline/.style={
        rectangle, draw=blue!60!black, fill=blue!10, thick, 
        minimum width=3.5cm, minimum height=1.2cm, 
        align=center, inner sep=5pt
    },
    process_online/.style={
        rectangle, draw=green!40!black, fill=green!10, thick, 
        minimum width=3.5cm, minimum height=1.2cm, 
        align=center, inner sep=5pt
    },
    arrow/.style={->, thick, draw=black!70, rounded corners=5pt}
]
\node[font=\Large\bfseries] (offTitle) {Offline Training};

\node[font=\Large\bfseries, right=2.3cm of offTitle] (onTitle) {Online Serving};

\node[data, below=1.2cm of onTitle, fill=gray!10] (query) {\textbf{User Query}};

\node[process_online, below=1.3cm of query, minimum height=1cm] (tok) {Tokenizer};
\node[process_online, below=0.8cm of tok, minimum height=1cm] (inf) {Inference};
\node[process_online, below=0.8cm of inf, minimum height=1cm] (pool) {Mean Pooling};
\node[process_online, below=0.8cm of pool, minimum height=1cm] (norm) {L2 Norm};

\begin{scope}[on background layer]
    \node[draw=green!60!black, thick, fill=green!5, fit=(tok) (norm), inner sep=10pt, label={[anchor=south, yshift=5pt]north:\textbf{Encoding Pipeline}}] (pipelineBox) {};
\end{scope}

\node[process_online, below=1.2cm of pipelineBox, fill=green!20] (faissOnline) {\textbf{FAISS Flat Index}\\(Brute Force)};
\node[process_online, below=1.0cm of faissOnline, fill=green!20] (metaOnline) {\textbf{Metadata Retrieval}\\(Cache)};
\node[data, below=1.0cm of metaOnline, fill=gray!10] (results) {\textbf{Ranked Results}};

\draw[arrow] (query) -- (pipelineBox.north);
\draw[arrow] (tok) -- (inf);
\draw[arrow] (inf) -- (pool);
\draw[arrow] (pool) -- (norm);
\draw[arrow] (pipelineBox.south) -- node[midway, right, font=\footnotesize]{Query Vector} (faissOnline.north);
\draw[arrow] (faissOnline) -- node[midway, right, font=\footnotesize]{Top-K IDs} (metaOnline);
\draw[arrow] (metaOnline) -- (results);

\node[process_offline] (indexing) at (offTitle |- faissOnline) {\textbf{Indexing}\\(FAISS Flat Index)};

\node[process_offline, above=1.0cm of indexing] (model) {\textbf{Fine-Tuned}\\\textbf{PyTorch Model}};
\node[process_offline, above=1.0cm of model] (finetune) {\textbf{Fine-Tuning}\\(SentenceTransformer)};
\node[process_offline, above=1.0cm of finetune] (preprocess) {\textbf{Preprocessing}\\\textbf{\& Tokenization}};

\node[process_offline, below=1.0cm of indexing] (lookup) {\textbf{Metadata Lookup}\\\textbf{(Text-based)}};

\node[process_offline, left=2.5cm of finetune, fill=blue!5, draw=blue!40, minimum width=2.8cm] (loss) {MNRL Loss};
\node[data, left=1.2cm of preprocess] (rawData) {\textbf{Raw Data}};
\node[process_offline, left=2.2cm of indexing] (vectorize) {\textbf{Vectorization}\\(GPU)};
\node[data, left=1.2cm of vectorize] (catalog) {\textbf{Catalog}};
\node[data, left=1.2cm of lookup] (metadata) {\textbf{Metadata}};

\draw[arrow] (rawData) -- (preprocess);
\draw[arrow] (preprocess) -- (finetune);
\draw[arrow] (finetune) -- node[midway, above, xshift= 4pt, font=\footnotesize]{Supervisor} (loss);
\draw[arrow] (finetune) -- (model);
\draw[arrow] (model) -- (indexing);
\draw[arrow] (catalog) -- (vectorize);
\draw[arrow] (vectorize) -- node[midway, above, font=\footnotesize]{Embeddings} (indexing);
\draw[arrow] (metadata) -- (lookup);
\draw[arrow] (indexing) -- (lookup);

\draw[arrow, green!40!black, ultra thick] (indexing.east) -- node[midway, above, font=\footnotesize, fill=white, inner sep=2pt]{Load Index} (faissOnline.west);

\end{tikzpicture}

%% file: fig03_architecture_optimized.tex
\begin{tikzpicture}[
    font=\sffamily\small,
    >=LaTeX, 
    node distance=1.0cm and 1.0cm, 
    data_node/.style={
        rectangle, draw=orange!80!black, fill=orange!10, thick, 
        minimum width=2.5cm, minimum height=1.2cm, 
        align=center, inner sep=5pt
    },
    off_proc/.style={
        rectangle, draw=blue!80!black, fill=blue!10, thick, 
        minimum width=3.5cm, minimum height=1.2cm, 
        align=center, inner sep=5pt, rounded corners=4pt
    },
    artifact/.style={
        rectangle, draw=gray!80!black, fill=gray!20, thick, 
        minimum width=3.0cm, minimum height=1.2cm, 
        align=center, inner sep=5pt, rounded corners=2pt, dashed
    },
    on_proc/.style={
        rectangle, draw=green!80!black, fill=green!10, thick, 
        minimum width=3.5cm, minimum height=1.2cm, 
        align=center, inner sep=5pt, rounded corners=4pt
    },
    arrow/.style={->, thick, draw=black!70, rounded corners=5pt},
    load_arrow/.style={->, ultra thick, draw=blue!60!black}
]

\node[data_node, fill=gray!10] (query) {\textbf{User Query}};

\node[on_proc, below=1.3cm of query] (infEnc) {Query Encoding\\(ONNX Runtime)};
\node[on_proc, below=1.3cm of infEnc] (infSearch) {Vector Search\\(FAISS HNSW)};
\node[on_proc, below=1.3cm of infSearch] (infLookup) {Metadata Enrichment\\(O(1) Lookup)};

\begin{scope}[on background layer]
    \node[draw=green!60!black, thick, fill=green!5, rounded corners=8pt, fit=(infEnc) (infSearch) (infLookup), inner sep=15pt, label={[anchor=south, font=\bfseries]north:Inference Service}] (serviceBox) {};
\end{scope}

\node[data_node, below=1.2cm of serviceBox, fill=green!20] (response) {JSON Response};

\draw[arrow] (query) -- (serviceBox.north);
\draw[arrow] (infEnc) -- (infSearch);
\draw[arrow] (infSearch) -- (infLookup);
\draw[arrow] (serviceBox.south) -- (response);


\node[artifact, left=4.5cm of infEnc] (artOnnx) {\textbf{ONNX Model}\\(.onnx, Int8)};
\node[artifact] (artFaiss) at (artOnnx |- infSearch) {\textbf{FAISS Index}\\(.index, HNSW)};
\node[artifact] (artPickle) at (artOnnx |- infLookup) {\textbf{Metadata Cache}\\(.pkl, Hash Map)};

\node[off_proc, above=1.0cm of artOnnx] (quantStep) {\textbf{ONNX Conversion}\\(Quantization)};
\node[off_proc, above=1.0cm of quantStep] (trainStep) {\textbf{Bi-Encoder Training}\\(PyTorch, MNRL)};
\node[data_node, above=1.0cm of trainStep] (trainData) {Training Pairs};

\node[off_proc] (vecStep) at ($(artOnnx)!0.5!(artFaiss)$) {\textbf{Vectorization}\\(Batch Inference)};

\node[off_proc, left=1.0cm of vecStep] (metaStep) {\textbf{Metadata Filtering}\\(Spark/Pandas)};
\node[data_node, above=1.0cm of metaStep] (metaData) {Raw Catalog};

\draw[arrow] (trainData) -- (trainStep);
\draw[arrow] (trainStep) -- (quantStep);
\draw[arrow] (quantStep) -- (artOnnx);
\draw[arrow] (artOnnx) -- (vecStep);
\draw[arrow] (vecStep) -- (artFaiss);

\draw[arrow] (metaData) -- (metaStep);
\draw[arrow] (metaStep) |- (artPickle);
\draw[arrow] (metaStep) -- (vecStep);

\node[font=\Large\bfseries, above=0.8cm of trainData] (offTitle) {Offline Optimization};

\node[font=\Large\bfseries] (onTitle) at (offTitle -| query) {Online Serving};

\draw[load_arrow] (artOnnx) -- node[midway, above, font=\footnotesize, text=blue!60!black]{Load Model} (infEnc);
\draw[load_arrow] (artFaiss) -- node[midway, above, font=\footnotesize, text=blue!60!black]{Load Index} (infSearch);
\draw[load_arrow] (artPickle) -- node[midway, above, font=\footnotesize, text=blue!60!black]{Load Cache} (infLookup);

\coordinate (sepTop) at ($(offTitle.east)!0.5!(onTitle.west)$);
\coordinate (sepBot) at (sepTop |- response.south);

\end{tikzpicture}

%% file: fig04_results.tex
\begin{tikzpicture}
\begin{axis}[
    ybar,
    bar width=12pt,
    width=0.65\linewidth,
    height=0.32\linewidth,
    enlarge x limits=0.25,
    ymin=0,
    ymax=0.75,
    ylabel={Metric value},
    ymajorgrids=true,
    grid style={dashed,gray!25},
    enlarge y limits={upper, value=0.06},
    symbolic x coords={BM25, Zero-Shot, Fine-Tuned},
    xtick=data,
    tick label style={font=\small},
    label style={font=\small},
    nodes near coords={\pgfmathprintnumber[fixed,precision=2]{\pgfplotspointmeta}},
    nodes near coords style={font=\scriptsize, yshift=2pt},
    legend columns=2,
    legend style={
        at={(0.5,1.06)},
        anchor=south,
        draw=none,
        fill=none,
        font=\small,
        /tikz/every even column/.append style={column sep=8pt}
    },
    legend cell align=left,
    clip=false
]

\addplot[fill=blue!60] coordinates {(BM25,0.26) (Zero-Shot,0.42) (Fine-Tuned,0.66)};
\addplot[fill=orange!70] coordinates {(BM25,0.25) (Zero-Shot,0.36) (Fine-Tuned,0.59)};

\legend{Recall\textsuperscript{@}10, MRR\textsuperscript{@}10}

\end{axis}
\end{tikzpicture}